\documentclass[preprint, 3p, authoryear]{elsarticle}

\usepackage[monochrome]{color}
\usepackage{amssymb}
\usepackage[utf8]{inputenc}
\usepackage{amsmath}

\usepackage{subcaption}
\usepackage{tikz}
\usetikzlibrary{positioning}
\usepackage{hyperref}

\definecolor{lightgreen}{RGB}{128, 255, 170}
\definecolor{lightblue}{RGB}{179, 209, 255}
\definecolor{lightorange}{RGB}{255, 230, 128}

\DeclareMathOperator*{\argmax}{arg\,max}

\begin{document}
\begin{frontmatter}
\title{Constructing a Natural Language Inference Dataset using Generative Neural Networks}
\author[ijs]{Janez Starc\corref{cor1}}
\ead{janez.starc@ijs.si}
\author[ijs]{Dunja Mladenić}
\ead{dunja.mladenic@ijs.si}

\address[ijs]{Jožef Stefan Institute and Jožef Stefan International Postgraduate School, Jamova 39, 1000 Ljubljana, Slovenia}
\cortext[cor1]{Corresponding author}
\begin{abstract}
Natural Language Inference is an important task for Natural Language Understanding. It is concerned with classifying the logical relation between two sentences. 
In this paper, we propose several text generative neural networks for generating text hypothesis, which allows construction of new  Natural Language Inference datasets.  
To evaluate the models, we propose a new metric -- the accuracy of the classifier trained on the generated dataset. The accuracy obtained by our best generative model is only 2.7\% lower than the accuracy of the classifier trained on the original, human crafted dataset. Furthermore, the best generated dataset combined with the original dataset achieves the highest accuracy. 
The best model learns a mapping embedding for each training example. 
By comparing various metrics we show that datasets that obtain higher ROUGE or METEOR scores do not necessarily yield higher classification accuracies.  
We also provide analysis of what are the characteristics of a good dataset including the distinguishability of the generated datasets from the original one.
\end{abstract}

\begin{keyword}
natural language inference \sep natural language generation \sep machine learning \sep dataset construction \sep generative neural network \sep recurrent neural network
\end{keyword}
\end{frontmatter}
\section{Introduction}
\label{sec:nli_intro}

The challenge in Natural Language Inference (NLI), also known as Recognizing Textual Entailment (RTE), is to correctly decide whether a sentence (referred to as a premise) entails or contradicts or is neutral in respect to another sentence (a hypothesis). This classification task requires various natural language comprehension skills.
In this paper, we are focused on the following natural language generation task based on NLI. 
Given the premise the goal is to generate a stream of hypotheses that comply with the label (\emph{entailment}, \emph{contradiction} or \emph{neutral}).
In addition to reading capabilities this task also requires language generation capabilities.
  
The Stanford Natural Language Inference (SNLI) Corpus \citep{snliemnlp2015} is a NLI dataset that contains over a half a million examples. The size of the dataset is sufficient to train powerful neural networks. Several successful classification neural networks have already been proposed \citep{rocktaschel2016reasoning, wang2015learning, cheng2016long, parikh2016decomposable}.
In this paper, we utilize SNLI to train generative neural networks. Each example in the dataset consist of two human-written sentences, a premise and a hypothesis, and a corresponding label that describes the relationship between them. Few examples are presented in Table~\ref{tab:entail_orig_ex}.

The proposed generative networks are trained to generate a hypothesis given a premise and a label, which allow us to construct new, unseen examples. 
Some generative models are build to generate a single optimal response given the input. Such models have been applied to machine translation \citep{sutskever2014sequence}, image caption generation\citep{xu2015show}, or dialogue systems \citep{serban2016building}. Another type of generative models are autoencoders that generate a stream of random samples from the original distribution. For instance, autoencoders have been used to generate text \citep{bowman2015generating,li2015hierarchical}, and images \citep{goodfellow2014generative}. In our setting we combine both approaches to generate a stream of random responses (hypotheses) that comply with the input (premise, label). 

\begin{table}
	\centering  	
	
	\begin{tabular} {lll}

		Premise & Hypothesis & Label  \\
		\hline
		A person throwing a yellow ball in the air.   & The ball sails through the air & entailment \\
		A person throwing a yellow ball in the air. & The person throws a square & contradiction \\
		A person throwing a yellow ball in the air.  & The ball is heavy  & neutral \\
		\hline

	\end{tabular}
	\caption{Three NLI examples from SNLI.}
	\label{tab:entail_orig_ex}
\end{table}

But what is a good stream of hypotheses? We argue that a good stream contains \emph{diverse}, \emph{comprehensible}, \emph{accurate} and \emph{non-trivial} hypotheses. A hypothesis is comprehensible if it is grammatical and semantically makes sense. It is accurate if it clearly expresses the relationship (signified by the label) with the premise. Finally, it is non-trivial if it is not trivial to determine the relationship (label) between the hypothesis and premise. For instance, given a premise ''A man drives a red car'' and label \emph{entailment}, the hypothesis ''A man drives a car'' is more trivial than ''A person is sitting in a red vehicle''. 

The next question is how to automatically measure the quality of generated hypotheses. One way is to use metrics that are standard in text generation tasks, for instance ROUGE\citep{lin2004rouge}, BLEU\citep{papineni2002bleu}, METEOR\citep{meteor2014}.
These metrics estimate the similarity between the generated text and the original reference text. In our task they can be used by comparing the generated and reference hypotheses with the same premise and label. The main issue of these metrics is that they penalize the diversity since they penalize the generated hypotheses that are dissimilar to the reference hypothesis. 
An alternative metric is to use a NLI classifier to test the generated hypothesis if the input label is correct in respect to the premise. A perfect classifier would not penalize diverse hypotheses and would reward accurate and (arguably to some degree) comprehensible hypotheses. However, it would not reward non-trivial hypotheses. 

Non-trivial examples are essential in a dataset for training a capable machine learning model. Furthermore, we make the following hypothesis.

\begin{quote}
	A good dataset for training a NLI classifier consists of a variety of \emph{accurate, non-trivial and comprehensible} examples.
\end{quote}

Based on this hypothesis, we propose the following approach for evaluation of generative models, which is also presented in Figure~\ref{fig:nli_approach}. First, the generative model is trained on the original training dataset. 
Then, the premise and label from an example in the original dataset are taken as the input to the generative model to generate a new \emph{random} hypothesis. The generated hypothesis is combined with the premise and the label to form a new unseen example. 
This is done for every example in the original dataset to construct a new dataset.
Next, a classifier is trained on the new dataset. Finally, the classifier is evaluated on the original test set.
The accuracy of the classifier is the proposed quality metric for the generative model. It can be compared to the accuracy of the classifier trained on the original training set and tested on the original test set. 

The generative models learn solely from the original training set to regenerate the dataset. 
Thus, the model learns the distribution of the original dataset. Furthermore, the generated dataset is just a random sample from the estimated distribution. 
To determine how well did the generative model learn the distribution, we observe how \emph{close} does the accuracy of the classifier trained on the generated dataset approach the accuracy of classifier trained on the original dataset.
 
Our flagship generative network \textsc{EmbedDecoder} works in a similar fashion as the encoder-decoder networks, where the encoder is used to transform the input into a low-dimensional latent representation, from which the decoder reconstructs the input. The difference is that \textsc{EmbedDecoder} consists only of the decoder, and the latent representation is learned as an embedding for each training example separately. In our models, the latent representation represents the mapping between the premise and the label on one side and the hypothesis on the other side. 

Our main contributions are i) a novel \emph{generative neural network}, which consist of the decoder that learns a mapping embedding for each training example separately, ii) a procedure for generating NLI datasets automatically, iii) and a novel \emph{evaluation metric} for NLI generative models -- the accuracy of the classifier trained on the generated dataset. 

In Section~\ref{sec:nli_related} we present the related work. In Section~\ref{sec:nli_models} the considered neural networks are presented. Besides the main generative networks, we also present classification and discriminative networks, which are used for evaluation. 
The results are presented in Section~\ref{sec:nli_results}, where the generative models are evaluated and compared. From the experiments we can see that the best dataset was generated by the attention-based model \textsc{EmbedDecoder}. The classifier on this dataset achieved accuracy of $78.5\%$, which is $2.7\%$ less than the accuracy achieved on the original dataset. We also investigate the influence of latent dimensionality on the performance, compare different evaluation metrics, and provide deeper insights of the generated datasets. The conclusion is presented in Section~\ref{sec:nli_conclusion}.  

\begin{center}
	\begin{figure}
		\centering
		\begin{tikzpicture}[auto, thick, align=center]
		\node [draw, text width=1.9cm,fill =lightblue](orgdata){Original Dataset};
		\node [draw, right = 3cm of orgdata, rounded corners, text width=1.9cm, fill =lightgreen](genmodel){Generative Model};
		
		\draw[<-] (orgdata) edge node{learns from} (genmodel);
		\node [draw, right = 3cm of genmodel, text width=1.9cm,fill =lightblue](gendata){Generated Dataset};
		\draw [->] (genmodel) edge node{generates} (gendata);
		
		\node [draw, below = 3cm of  orgdata, text width=1.9cm, fill =lightgreen](orgclass) {Original Classifier};
		\draw [<-] (orgdata) edge node [text width = 2cm]{learns from, tested on} (orgclass);
		
		\node [draw, below = 3cm of gendata, rounded corners, text width=1.9cm, fill =lightgreen](genclass){New Classifier};
		\draw [<-] (gendata) edge node {learns from} (genclass);
		\draw [<-] (orgdata) edge node {tested on} (genclass);
		\draw [<->, dashed] (orgclass) edge node{accuracy compared} (genclass) ;
		
		\end{tikzpicture}
		
		\caption{Evaluation of NLI generative models. Note that both datasets are split on training test and validation sets.}
		\label{fig:nli_approach}
	\end{figure}
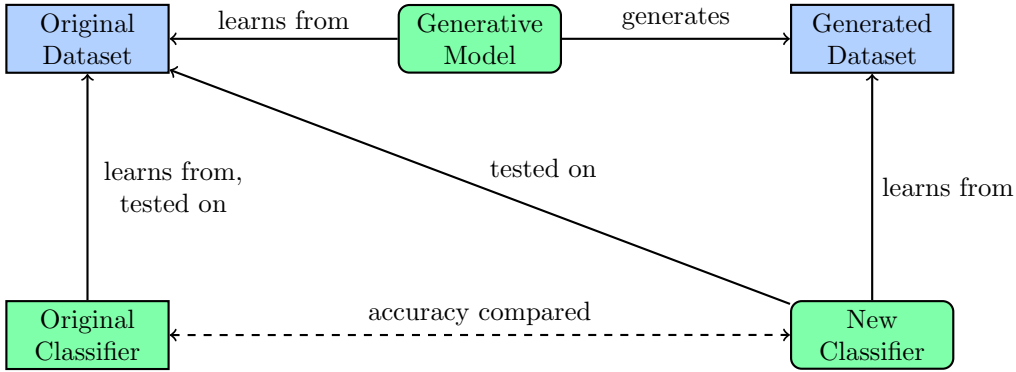
	
\end{center}

\section{Related Work}
\label{sec:nli_related}

NLI has been the focal point of Recognizing Textual Entailment (RTE) Challenges, where the goal is to determine if the premise entails the hypothesis or not\footnote{The task is designed as a 2-class classification problem, while the SNLI dataset is constructed for a 3-class classification task.}. 
The proposed approaches for RTE include bag-of-words matching approach \citep{glickman2005web}, matching predicate argument structure approach \citep{maccartney2006learning} and logical inference approach \citep{bos2006logical, tatu2006logic}. Another rule-based inference approach was proposed by \citet{barhaim2015}. This approach allows generation of new hypotheses by transforming parse trees of the premise while maintaining entailment. \citet{hickl2006recognizing} proposes an approach for constructing training datasets by extracting sentences from news articles that tend to be in an entailment relationship.

After SNLI dataset was released several neural network approaches for NLI classification have emerged. \citep{rocktaschel2016reasoning, wang2015learning, cheng2016long, parikh2016decomposable}. The state-of-the-art model \citep{parikh2016decomposable} achieves $86.6\%$ accuracy on the SNLI dataset. A similar generation approach to ours was proposed by \citet{kolesnyk2016generating}, The goal of this work is generating entailment inference chains, where only examples with entailment label are used. 

Natural Lanuguage Generation (NLG) is a task of generating natural language from a structured form such as knowledge base or logic form \citep{wen2015semantically, mairesse2010phrase, belz2008automatic}. The input in our task is unstructured text (premise) and label. On the other side of this spectrum, there are tasks that deal solely with unstructured text, like machine translation \citep{koehn2009statistical, bahdanau2014neural, luong2014addressing}, summarization \citep{clarke2008global, rush2015neural} and conversational dialogue systems \citep{serban2016building, banchs2012iris}. Another recently popular task is generating captions from images \citep{vinyals2015show, socher2014grounded}. 

With the advancement of deep learning, many neural network approaches have been introduced for generating sequences. 
The Recurrent Neural Network Language Model (RNNLM) \citep{mikolov2010recurrent} is one of the simplest neural architectures for generating text. The approach was extended by \citep{sutskever2014sequence}, which use encoder-decoder architecture to generate a sequence from the input sequence. The Hierarchical Recurrent Encoder-Decoder (HRED) architecture \citep{serban2016building} generates sequences from several input sequences. 
These models offer very little variety of output sequences. It is obtained by modeling the output distribution of the language model.
To introduce more variety, models based on variational autoencoder (VAE)\citep{kingma2013auto} have been proposed.
These models use stochastic random variables as a source of variety. In \citep{bowman2015generating} a latent variable is used to initial the RNN that generates sentences, while the variational recurrent neural network (VRNN) \citep{chung2015recurrent} models the dependencies between latent variables across subsequent steps of RNN. The Latent Variable Hierarchical Recurrent Encoder-Decoder (VHRED) \citep{serban2016hierarchical} extends the HRED by incorporating latent variables, which are learned similarly than in VAE. The latent variables are, like in some of our models, used to represent the mappings between sequences.
Conditional variational autoencoders (CVAEs) \citep{yan2015attribute2image} were used to generate images from continuous visual attributes. These attributes are conditional information that is fed to the models, like the discrete label is in our models.

As recognized by \citep{reiter2009investigation}, the evaluation metrics of text-generating models fall into three categories: manual evaluation, automatic evaluation metrics, task-based evaluation. In evaluation based on human judgment each generated textual example is inspected manually. The automatic evaluation metrics, like ROUGE, BLEU and METEOR, compare human texts and generated texts. 
\citep{elliott2014comparing} shows METEOR has the strongest correlation with human judgments in image description evaluation. 
The last category is task-based evaluation, where the impact of the generated texts on a particular task is measured. 
This type of evaluation usually involves costly and lengthy human involvement, like measuring the effectiveness of smoking-cessation letters \citep{reiter2003lessons}.
On the other hand, the task in our evaluation, the NLI classification, is automatic. In \citep{hodosh2013framing} ranking was used as an automatic task-based evaluation for associating images with captions.

\section{Models}
\label{sec:nli_models}
In this section, we present several neural networks used in the experiments. We start with variants of Recurrent Neural Networks, which are essential layers in all our models. Then, we present classification networks, which are needed in evaluation of generative neural networks presented in the following section. Next, we present how to use generative networks to generate hypothesis. Finally, we present discriminative networks, which are used for evaluation and analysis of the hypotheses.   

The premise $W^p = w^p_1 w^p_2 \ldots w^p_M$ and hypothesis $W^h = w^h_1 w^h_2 \ldots w^h_N$ are represented with word embeddings  $X^p = x^p_1 x^p_2 \ldots x^p_M$ and $X^h = x^h_1 x^h_2 \ldots x^h_N$ respectively. Each $x$ is a $e$-dimensional vector that represents the corresponding word, $M$ is the length of premise, and $N$ is the length of hypothesis. The labels (entailment, contradiction, neutral) are represented by a $3$-dimensional vector $Y^l$ if the label is the output of the model, or $L$ if the label is the input to the model.

\subsection{Recurrent Neural Networks}

The Recurrent Neural Networks (RNNs) are neural networks suitable for processing sequences. They are the basic building block in all our networks. We use two variants of RNNs --  Long short term memory (LSTM) network \citep{hochreiter1997long} and an attention-based extension of LSTM, the mLSTM \citep{wang2015learning}.
The LSTM tends to learn long-term dependencies better than vanilla RNNs.
The input to the LSTM is a sequence of vectors $X=x_1x_2 \ldots x_n$, and the output is a sequence of vectors $H=h_1h_2 \ldots h_n$. At each time point $t$, input gate $i_t$, forget gate $f_t$, output gate $o_t$, cell state $C_t$ and one output vector $h_t$ are calculated.

\begin{align}
i_t &= \sigma(W_i x_t + U_i h_{t-1} + b_i) \\ 
f_t &= \sigma(W_f x_t + U_f h_{t-1} + b_f) \\
o_t &= \sigma(W_o x_t + U_o h_{t-1} + b_o) \\
C_t &= f_t \odot C_{t-1} + i_t \odot tanh(W_c x_t + U_c h_{t-1} + b_c) \\
h_t &= o_t \odot tanh(C_t),
\end{align}

where $\sigma$ is a sigmoid function, $\odot$ is the element-wise multiplication operator, $W \in \mathbb{R}^{d \times e} $ and $U \in \mathbb{R}^{d \times d}  
$ are parameter matrices, $b \in \mathbb{R}^{d} $ parameter vectors, $e$ is the input vector dimension, and $d$ is the output vector dimension. 
The vectors $C_0$ and $h_0$ are set to zero in the standard setting, however, in some cases in our models, they are set to a value that is the result of previous layers.

The mLSTM is an attention-based model with two input sequences -- premise and hypothesis in case of NLI. Each word of the premise is matched against each word of the hypothesis to find the soft alignment between the sentences. 
The mLSTM is based on LSTM in such a way that it remembers the important matches and forgets the less important. 
The input to the LSTM inside the mLSTM at each time step is $x_t^\prime = [a_t, x_t^h]$, where $a_t$ is an attention vector that represents the weighted sum of premise sequence, where the weights present the degree to which each token of the premise is aligned with the $t$-th token of the hypothesis $x_t^h$, and $[*,*]$ is the concatenation operator. 
More details about mLSTM are presented in \citep{wang2015learning}.

\subsection{Classification model}
\label{sec:classifier}
The classification model predicts the label of the example given the premise and the hypothesis. 
We use the mLSTM-based model proposed by \citep{wang2015learning}. 

The architecture of the model is presented in Figure~\ref{fig:class_model}.
The embeddings of the premise $X^p$ and hypothesis $X^h$ are the input to the first two LSTMs to obtain the hidden states of the premise $H^p$ and hypothesis $H^h$. 

\begin{equation}
   H^p = \mathit{LSTM}(X^p) \quad \quad
   H^h = \mathit{LSTM}(X^h)
\end{equation}

All the hidden states in our models are $d$-dimensional unless otherwise noted. The hidden states $H^p$ and $H^h$ are the input to the mLSTM layer. The output of mLSTM are hidden states $H^m$, although only the last state $h^m_N$ is further used. A fully connected layer transforms it into a 3-dimensional vector, on top of which softmax function is applied to obtain the probabilities $Y^l$ of labels.        

\begin{equation}
Y^l = Softmax(\mathit{Dense}_3(h^m_N)),
\end{equation}
where $\mathit{Dense}_x$ represents the fully connected layer, whose output size is $x$.

\begin{center}
	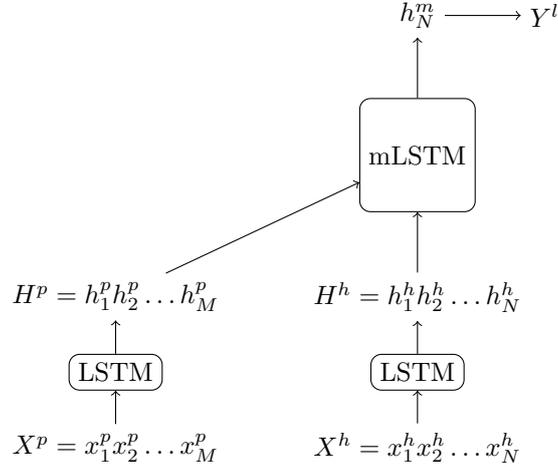
\begin{figure}
		\centering
		\begin{tikzpicture}
		\node (premise){$X^p = x^p_1 x^p_2 \ldots x^p_M$};
		\node [draw, above of = premise, rounded corners](lstmp){LSTM};
		\draw [->] (premise) edge (lstmp);
		\node [above of = lstmp](hpremise){$H^p = h^p_1 h^p_2 \ldots h^p_M$};
		\draw [->] (lstmp) edge (hpremise);
		
		\node [right=1cm of premise](hypo) {$X^h = x^h_1 x^h_2 \ldots x^h_N$};
		\node [draw, above of = hypo, rounded corners](lstmh){LSTM};
		\draw [->] (hypo) edge (lstmh);
		\node [above of = lstmh](hhypo){$H^h = h^h_1 h^h_2 \ldots h^h_N$};
		\draw [->] (lstmh) edge (hhypo);
		
		\node[draw, above =0.8cm of hhypo, minimum width=1.5cm,minimum height=1.5cm, rounded corners](attention){mLSTM};
		\draw [->] (hhypo) edge (attention);
		\draw [->] (hpremise) edge (attention);
		
		\node [above=0.8cm of attention](hiddenm) {$h^m_N$};
		\draw [->] (attention) edge (hiddenm);
		
		\node[right = 1cm of hiddenm](y){$Y^l$};
		\draw [->] (hiddenm) edge (y);
		
		\end{tikzpicture}
		
		\caption{NLI classification model}
        \label{fig:class_model}
	\end{figure}
\end{center}

\subsection{Generative models}
\label{sec:gen_models}
The goal of the proposed generative models, is to generate a diverse stream of hypotheses given the premise and the label. In this section, we present four variants of generative models, two variants of \textsc{EmbedDecoder} model presented in Figure~\ref{fig:embed_decoder}, and two variants of \textsc{EncoderDecoder} model presented in Figure~\ref{fig:encoder_decoder}.

\begin{center}
	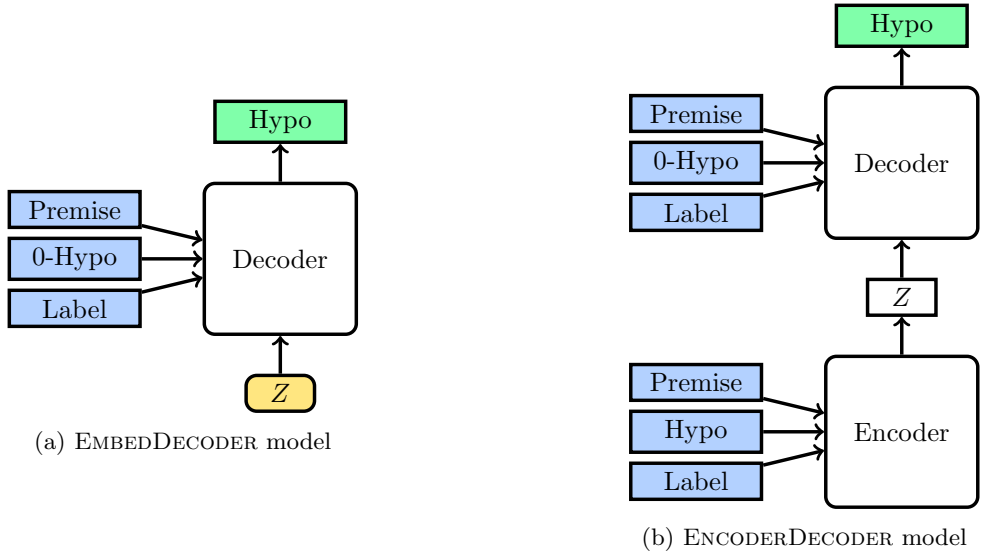
\begin{figure}
		\tikzset{every picture/.style={line width = 1.3pt}}
		\centering
		\begin{subfigure}{.49\textwidth}
				\centering
		\begin{tikzpicture}
		\node [draw, fill = lightorange, minimum width = 0.9cm, rounded corners](latent){$Z$};
		\node [draw, above = 0.5cm of latent, minimum size=2cm, rounded corners](decoder){Decoder};
		\node [draw, left = 0.8cm of decoder,  minimum width = 1.72cm, fill =lightblue](hypo){0-Hypo};
		\node [draw, above = 0.1cm of  hypo, minimum width = 1.72cm, fill =lightblue](premise){Premise};
		\node [draw, below = 0.1cm of hypo,  minimum width = 1.72cm, fill =lightblue](label){Label};
		\node [draw, above = 0.5cm of decoder, fill =lightgreen, minimum width = 1.72cm](output){Hypo};
		
		\draw [->] (latent) edge (decoder);
		\draw [->] (hypo) edge (decoder);
		\draw [->] (premise) edge (decoder);
		\draw [->] (label) edge (decoder);
		\draw [->] (decoder) edge (output);

		\end{tikzpicture}
        \caption{\textsc{EmbedDecoder} model}
        \label{fig:embed_decoder}
		\end{subfigure}
			\begin{subfigure}{.49\textwidth}
				\centering
				\begin{tikzpicture}
				\node [draw, minimum width = 0.9cm](latent){$Z$};
				\node [draw, above = 0.5cm of latent, minimum size=2cm, rounded corners](decoder){Decoder};
				\node [draw, left = 0.8cm of decoder,  minimum width = 1.72cm, fill =lightblue](hypo){0-Hypo};
				\node [draw, above = 0.1cm of  hypo, minimum width = 1.72cm, fill =lightblue](premise){Premise};
				\node [draw, below = 0.1cm of hypo,  minimum width = 1.72cm, fill =lightblue](label){Label};
				\node [draw, above = 0.5cm of decoder, fill =lightgreen, minimum width = 1.72cm](output){Hypo};
				
				\draw [->] (latent) edge (decoder);
				\draw [->] (hypo) edge (decoder);
				\draw [->] (premise) edge (decoder);
				\draw [->] (label) edge (decoder);
				\draw [->] (decoder) edge (output);

				\node [draw, below = 0.5cm of latent, minimum size=2cm, rounded corners](encoder){Encoder};
				\node [draw, left = 0.8cm of encoder,  minimum width = 1.72cm, fill =lightblue](hypo2){Hypo};
				\node [draw, above = 0.1cm of  hypo2, minimum width = 1.72cm, fill =lightblue](premise2){Premise};
				\node [draw, below = 0.1cm of hypo2,  minimum width = 1.72cm, fill =lightblue](label2){Label};

				\draw [->] (encoder) edge (latent);
				\draw [->] (hypo2) edge (encoder);
				\draw [->] (premise2) edge (encoder);
				\draw [->] (label2) edge (encoder);

				\end{tikzpicture}
				
				\caption{\textsc{EncoderDecoder} model}
                \label{fig:encoder_decoder}
			\end{subfigure}
		\label{fig:gen_model_architecture}
		\caption{Generative models architecture. The rounded boxes represent trainable parameters, blue boxes are inputs, green boxes are outputs and the orange box represents the mapping embeddings. \emph{0-Hypo} denotes the shifted \footnotesize{$<$null$>$}-started hypothesis. Note that in \textsc{EncoderDecoder} model the latent representation $Z$ is just a hidden layer, while in \textsc{EmebedDecoder} it is a trainable parameter matrix.}
	\end{figure}
\end{center}

All models learn a latent representation $Z$ that represents the mapping between the premise and the label on one side, and the hypothesis on the other side.
The \textsc{EmbedDecoder} models learn the latent representation by learning an embedding of the mapping for each training example separately. The embedding for $i$-th training example $Z^{(i)}$ is a $z$-dimensional trainable parameter vector. Consequentely, $Z \in \mathbb{R}^{n \times z}$ is a parameter matrix of all embeddings, where $n$ is the number of training examples. On the other hand, in \textsc{EncoderDecoder} models latent representation is the output of the decoder.

The \textsc{EmbedDecoder} models are trained to predict the next word of the hypothesis given the previous words of hypothesis, the premise, the label, and the latent representation of the example.
\begin{equation}
\theta^\star, Z^\star = \argmax_{\theta, Z} \sum_{i=1}^n \sum_{k=1}^{d(W^{h^{(i)}})} \log p(w_k^{h^{(i)}} | w_{k-1}^{h^{(i)}}\ldots w_1^{h^{(i)}}, W^{p^{(i)}}, L^{(i)}, Z^{(i)}, \theta )
\end{equation}
where $\theta$ represent parameters other than $Z$, and $d(W^{h^{(i)}})$ is the length of the hypothesis $W^{h^{(i)}}$.
	
The \textsc{AttEmbedDecoder}, presented in Figure~\ref{fig:gen_model}, is attention based variant of \textsc{EmbedDecoder}. 	
The same mLSTM layer is used as in classification model. However, the initial cell state $C_0$ of mLSTM is constructed from the latent vector and the label input.
\begin{equation}
C_0 = \mathit{Dense}_d([Z^{(i)}, L])
\label{eq:reduction}
\end{equation}
For the sake of simplifying the notation, we dropped the superscript $(i)$ from the equations, except in $Z^{(i)}$, where we explicitly want to state that the embedding vector is used. 

The premise and the hypothesis are first processed by LSTM and then fed into the mLSTM, like in the classification model, however here the hypothesis is shifted. The first word of the hypothesis input is an empty token {\footnotesize{$<$null$>$}}, symbolizing the empty input sequence when predicting the first word. The output of the mLSTM is a hidden state $H^m$, where each $h^m$ represents an output word. To obtain the probabilities for all the words in the vocabulary $y_k^h$ for the position $k$ in the output sequence, $h_k^m$ is first transformed into a vocabulary-sized vector, then the softmax function is applied.
\begin{equation}
y_k^h = softmax(\mathit{Dense}_V(h_k^m)),
\end{equation} 
where V is the size of the vocabulary. But, due to the large size of the vocabulary, a two-level hierarchical softmax \citep{goodman2001classes} was used instead of a regular softmax to reduce the number of parameters updated during each training step.
\begin{equation}
y_k^h = hsoftmax(h_k^m)
\end{equation}
In the training step, the last output word $y^h_{N+1}$ is set to {\footnotesize{$<$null$>$}}, while in the generating step, it is ignored.

In the \textsc{EmbedDecoder} model without attention, \textsc{BaseEmbedDecoder}, the mLSTM is replaced by a regular LSTM. The input to this LSTM is the shifted hypothesis. But, here the premise is provided through the initial cell state $C_0$. Specifically, last hidden state of the premise is merged with class input and the latent representation, then fed to the LSTM. 

\begin{equation}
C_0 = \mathit{Dense}_{d^\prime}([Z^{(i)},L, h_M^p])
\label{eq:reduction2}
\end{equation}
In order to not lose information $d^\prime$ was picked to be equal to sum of the sizes of $Z^{(i)}$, $L$ and $h_M^p$. Thus, $d^\prime = f + 3 + d$. Since the size of $C_0$ is $d^\prime$, the output vectors of the LSTM are also the size of $d^\prime$. 

We also present two variants of \textsc{EncoderDecoder} models, a regular one \textsc{BaseEncodeDecoder}, and a regularized one \textsc{VarEncoderDecoder}, which is based on Variational Bayesian approach. 
As presented in Figure~\ref{fig:encoder_decoder}, all the information (premise, hypothesis, label) is available to the encoder, whose output is the latent representation $Z$. On the other hand, the decoder is provided with the same premise and label, but the hypothesis is shifted. 
This forces the encoder to learn to encode only the missing information -- the mapping between premise-label pair and the hypothesis.
The encoder has a similar structure as the classification model in Figure~\ref{fig:class_model}. 
Except that the label is connected to the initial cell state of the mLSTM

\begin{equation}
C_0 = \mathit{Dense}_d(L),
\end{equation}
and the output of mLSTM $h_N^m$ is transformed into latent representation $Z$

\begin{equation}
Z =  \mathit{Dense}_z(h_N^m).
\end{equation}
The decoder is the same as in \textsc{EmbedDecoder}. 

The \textsc{VarEncoderDecoder} models is based on Variational Autoencoder from \citep{kingma2013auto}.
Instead of using single points for latent representation as in all previous models, the latent representation in \textsc{VarEncoderDecoder} is presented as a continuous variable $Z \sim  \mathcal{N}(Z_\mu, Z_\sigma)$. Thus, the mappings are presented as a soft elliptical regions in the latent space, instead of a single points, which forces the model to fill up the latent space \citep{bowman2015generating}. Both $Z_\mu$ and $Z_\sigma$ are calculated form the output of the encoder using two different fully connected layers.

\begin{align*}
Z_\mu = \mathit{Dense}_z(h_N^m), \quad Z_\sigma= \mathit{Dense}_z(h_N^m).
\end{align*}

To sample from the distribution the reparametrization trick is applied

\begin{equation}
Z = Z_\mu + Z_\sigma \odot \epsilon, \quad \epsilon \sim  \mathcal{N}(0, \mathbb{I})
\end{equation}
When training, a single sample is generated per example to generate $Z$.

As in \citep{kingma2013auto}, the following regularization term is added to the loss function 
\begin{equation}
\frac{1}{2} ( 1 + log(Z_\sigma^2) - Z_\mu^2 - Z_\sigma^2).
\end{equation}

\subsection{Generating hypotheses}

In the generation phase only decoder of a trained generative model is used. It generates a hypothesis given the premise, label, and a randomly selected latent vector $Z^{(*)}$ .
A single word is generated in each step, and it becomes the hypothesis input in the next step. 

\begin{equation} 
x^h_k = \mathit{embedding}(\argmax{y_k^h})
\label{eq:max_y}
\end{equation}
 
We also used beam search to optimize hypothesis generation. 
Similarly as in \citep{sutskever2014sequence}, a small number of hypotheses are generated given a single input, then the best is selected. 
In $k$-beam search, in each time step  $k$ best partial hypotheses are expanded by all the words in the vocabulary producing $kV$ partial hypothesis. Out of these $k$ best partial hypotheses are selected for the next step according to the joint probability of each partial hypothesis. Thus, when $k$ is $1$, the procedure is the same as the one presented in Eq~\ref{eq:max_y}. 
The generation ends when {\footnotesize{$<$null$>$}} symbol is encountered or maximum hypothesis length is reached\footnote{In beam search mode the process stops when all $k$ hypotheses reach the {\footnotesize{$<$null$>$}} symbol or maximum hypothesis length is reached}.
The random latent vector $Z^{(*)}$ is selected randomly from a normal distribution $\mathcal{N}(0, \sigma)$, where $\sigma$ is the standard deviation of $Z$.

\begin{center}
	\begin{figure}
		\centering
		\begin{tikzpicture}
		\node (premise){$X^p = x^p_1 x^p_2 \ldots x^p_M$};
		\node [draw, above of = premise, rounded corners](lstmp){LSTM};
		\draw [->] (premise) edge (lstmp);
		\node [above of = lstmp](hpremise){$H^p = h^p_1 h^p_2 \ldots h^p_M$};
		\draw [->] (lstmp) edge (hpremise);
		
		\node [right=1cm of premise](hypo) {$X^h = \text{\footnotesize{$<$null$>$}} x^h_1 x^h_2 \ldots x^h_N$};
		\node [draw, above of = hypo, rounded corners](lstmh){LSTM};
		\draw [->] (hypo) edge (lstmh);
		\node [above of = lstmh](hhypo){$H^h = h^h_1 h^h_2 \ldots h^h_{N+1}$};
		\draw [->] (lstmh) edge (hhypo);
		
		\node[draw, above =0.8cm of hhypo, minimum width=1.5cm,minimum height=1.5cm, rounded corners](attention){mLSTM};
		\draw [->] (hhypo) edge (attention);
		\draw [->] (hpremise) edge (attention);

		\node [draw, circle, left =0.8cm of attention](sum){.};
		\node [above left of = sum](latent){$Z^{(i)}$};
		\node [below left of = sum](classinput){$X^l$};
		
		\draw [->] (latent) edge (sum);
		\draw [->] (classinput) edge (sum);
		\draw [->] (sum) edge (attention);
		
		\node [above=0.8cm of attention](hiddenm) {$H^m = h^m_1 x^m_2 \ldots h^m_{N+1}$};
		\draw [->] (attention) edge (hiddenm);
		
		\node[draw, above of = hiddenm, rounded corners](hsoftmax){Hierachical Softmax};
		\draw [->] (hiddenm) edge (hsoftmax);
		
		\node[above of = hsoftmax](output){$Y^h = y^h_1 y^h_2 \ldots y^h_{N+1}$};

		\end{tikzpicture}
		\caption{\textsc{AttEmbedDecoder} model}
		\label{fig:gen_model}
		
	\end{figure}
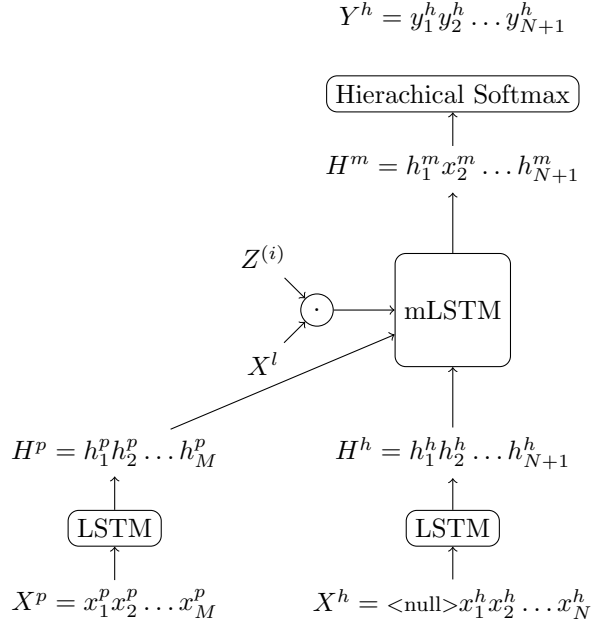
\end{center} 

\subsection{Discriminative model}
The discriminative model is used to measure the distinguishability between the original human written sentences and the generated ones. 
Higher error rate of the model means that the generative distribution is similar to the original distribution, which is one of the goals on the generative model.   
The model is based on Generative Adversarial Nets \citep{goodfellow2014generative}, where in a single network the generative part tires to trick the discriminative part by generating images that are similar to the original images, and the discriminative part tries to distinguish between the original and generated images. 
Due to the discreteness of words (the output of our generative model) it is difficult to connect both the discriminative and generative part in a single differentiable network, thus we construct them separately. 
The generative models have already been defined in Section~\ref{sec:gen_models}. Here we define the discriminative model. 

The discriminative model $D$ takes sequence $X$ and process it with LSTM and fully connected layer

\begin{equation}
D(X) = \sigma(\mathit{Dense}_1(\mathit{LSTM}(X))
\end{equation}
In the training step, one original sequence $X_\mathit{original}$ and one generated sequence $X_\mathit{generated}$ are processed by the discriminative model. The optimization function maximizes the following objective
\begin{equation}
log(D(X_\mathit{original})) + log(1 - D(X_\mathit{generated}))
\end{equation}
In the testing step, the discriminative model predicts correctly if 
\begin{equation}
D(X_\mathit{original}) > D(X_\mathit{generated})
\end{equation}

\section{Dataset Generation}

To construct a new dataset, first a generative model is trained on the training set of the original dataset. 
Then, a new dataset is constructed by generating a new hypotheses with a generative model. The premises and labels from the examples of the original dataset are taken as an input for the generative model. The new hypotheses replace the training hypotheses in the new dataset. 

Next, the classifier, presented in Section~\ref{sec:classifier}, is trained on the generated dataset. 
The accuracy of the new classifier is the main metric for evaluating the quality of the generated dataset.

\subsection{Experiment details}

All the experiments are performed on the SNLI dataset. 
There are 549,367 examples in the dataset, divided into training, development and test set. 
Both the development and test set contain around 10.000 examples.
Some examples are labeled with '\emph{-'}, which means there was not enough consensus on them. These examples are excluded.
Also, to speed up the computation we excluded examples, which have the premise longer than 25 words, or the hypothesis longer than 15 words. 
There were still $92.5\%$ remaining examples. Both premises and hypothesis were padded with {\footnotesize{$<$null$>$}} symbols (empty words), so that all premises consisted of 25 words, and all hypotheses consisted of 15 tokens. 

We use 50-dimensional word vectors\footnote{http://nlp.stanford.edu/data/glove.6B.zip} trained with GloVe \citep{pennington2014glove}. 
For words without pretrained embeddings, the embeddings are randomly selected from the normal distribution. 
Word embeddings are not updated during training.

For optimization Adam method \citep{kingma2014adam} was used with suggested hyperparameters\footnote{As suggested in \citep{kingma2014adam} $\beta_1$ is set to 0.9 and $beta_2$ is set to 0.999.}.

Classification models are trained until the loss on the validation set does not improve for three epochs.
The model with best validation loss is retained.

Generative models are trained for 20 epochs, since it turned out that none of the stopping criteria were useful.
With each generative model a new dataset is created. The new dataset consists of training set, which is generated using examples from the original training set, and a development set, which is generated from the original development set. The beam size for beam search was set to 1. The details of the decision are presented in Section~\ref{sec:ld_eval}. 

Some datasets were constructed by filtering the generated datasets according to various thresholds. Thus, the generated datasets were constructed to contain enough examples, so that the filtered datasets had at least the number of examples as the original dataset. In the end, all the datasets were trimmed down to the size of the original dataset by selecting the samples sequentially from the beginning until the dataset had the right size. Also, the datasets were filtered so that each of the labels was represented equally.  
All the models, including classification and discriminative models, were trained with hidden dimension $d$ set to 150, unless otherwise noted.

Our implementation is accessible at \url{http://github.com/jstarc/nli_generation}. It is based on libraries \emph{Keras}\footnote{\url{http://keras.io}} and \emph{Theano}\citep{theano}.

\section{Results}
\label{sec:nli_results}

First, the classification model \textsc{OrigClass} was trained on the original dataset.
This model was then used throughout the experiments for filtering the datasets, comparison, etc. 
Notice that we have assumed \textsc{OrigClass} to be ground truth for the purpose of our experiments.
However, the accuracy of this model on the original test set was $81.3\%$, which is less than $86.1\%$, which was attained by \emph{mLSTM (d=150)} model in \citep{wang2015learning}. Both models are very similar, including the experimental settings, however ours was trained and evaluated on a slightly smaller dataset. 

\subsection{Preliminary evaluation}
\label{sec:ld_eval}
Several \textsc{AttEmbedDecoder} models with various latent dimensions $z \in [2, 4, 8, 16, 32, 147\footnote{Latent dimension $z=147$ is the largest dimension so that there is no reduction in dimensionality in Equation~\ref{eq:reduction}, therefore $z + c = d$, where $c$ is the number of labels.}]$ were first trained and then used to generate new datasets. 
A couple of generated examples are presented in Table~\ref{tab:enatil_ex}.

\begin{table}
	\centering  	
	
	\begin{tabular} {|c|l|l|}
		\hline
		\multicolumn{2}{|c|}{Premise}	& A person throwing a yellow ball in the air. \\
		\hline
		$z=2$ & neutral & Someone is playing basketball. \\
		& contradiction & A person is sleeping in a chair. \\
		& entailment & A person is throwing a ball \\
		\hline
		$z=8$ & neutral & The person has a yellow ball going to the game. \\
		& contradiction & The person is sitting in the bleachers. \\
		& entailment & A person is playing with a ball. \\
		\hline
		$z=147$ & neutral & A person is trying to get home from give a ball. \\
		& contradiction & A person is reading a bank from london. \\
		& entailment & A person is throwing a ball up. \\
		\hline
		\noalign{\bigskip}
		\hline
		\multicolumn{2}{|c|}{Premise}	& Two women in bathing suits climb rock piles by the ocean. \\
		\hline
		$z=2$ & neutral & Two women are climbing rocks in the ocean on a sunny day. \\
		& contradiction & The women are playing basketball. \\
		& entailment & Two women are climbing. \\
		\hline
		$z=8$ & neutral & Two young women in bathing suits are friends \\
		& contradiction & Two women naked. \\
		& entailment & The girls looking at the water. \\
		\hline
		$z=147$ & neutral & Two women are looking at the lagoon in front of a calm shore. \\
		& contradiction & Two women are gossiping on a sandy beach. \\
		& entailment & A group of women are climbing wood in the ocean. \\
		\hline
	\end{tabular}
    \caption{Generated examples to illustrate the proposed appraoch.}
	\label{tab:enatil_ex}
\end{table}

\begin{figure}
	\centering
	\includegraphics[width=0.7\textwidth]{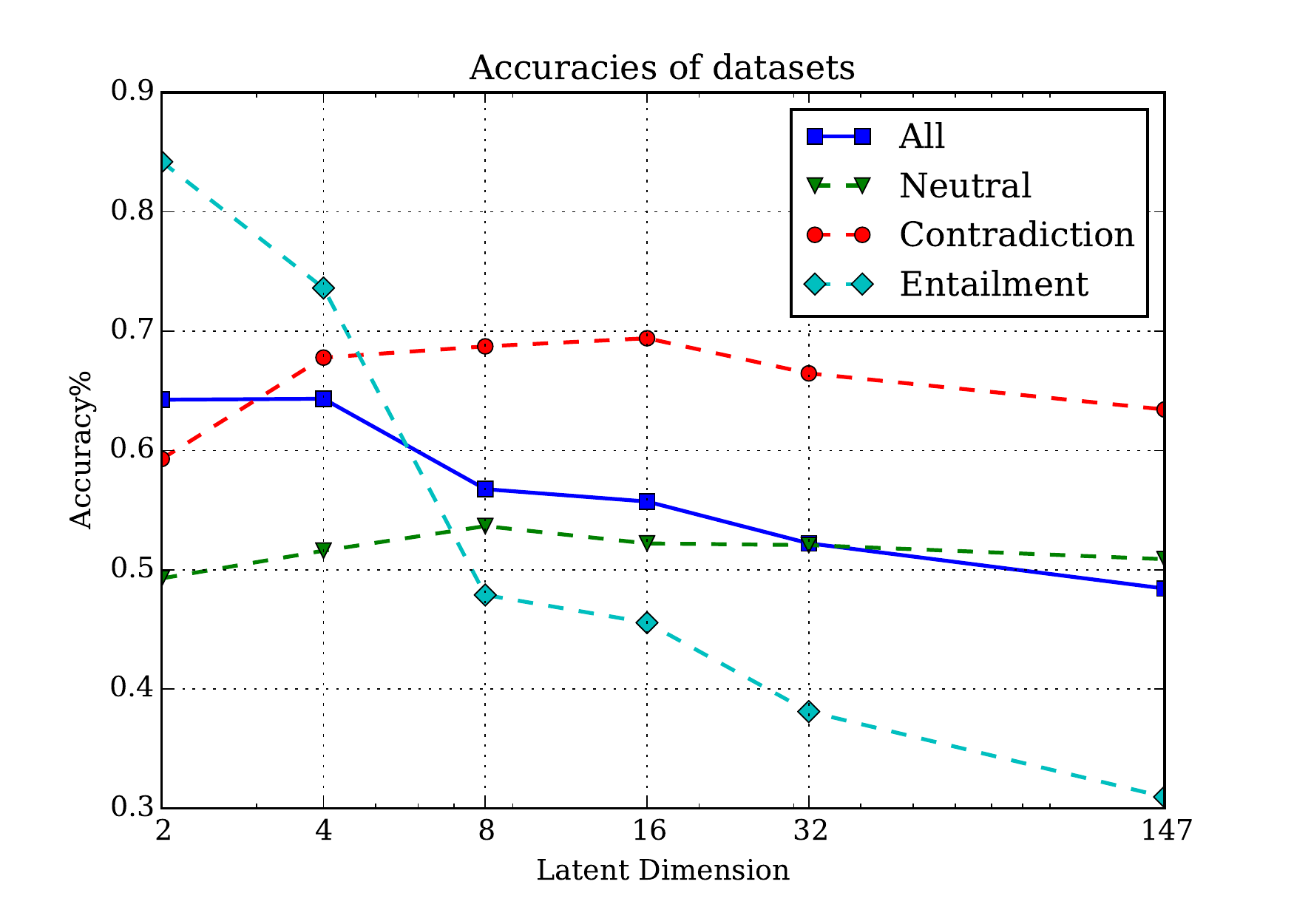}
	\caption{Accuracies of the unfiltered generated datasets classified by \textsc{OrigClass}. A dataset was generated for each generative model with different latent dimension $z \in [2, 4, 8, 16, 32, 147]$. For each dataset the examples were classified with \textsc{OrigClass}. The predicted labels were taken as a golden truth and were compared to the labels of the generated dataset to measure its accuracy. The accuracies were measured for all the labels together and for each label separately.}
	\label{fig:data_acc}
\end{figure}

Figure~\ref{fig:data_acc} shows the accuracies of the generated development datasets evaluated by the \textsc{OrigClass}. The maximum accuracy of $64.2\%$ was achieved by \textsc{EmbedDecoder} (z=2), and the accuracy is decreasing with the number of dimensions in the latent variable.
The analysis for each label shows that the accuracy of contradiction and neutral labels is quite stable, while the accuracy of the entailment examples drops significantly with latent dimensionality. One reason for this is that the hypothesis space of the entailment label is smaller than the spaces of other two labels. Thus, when the dimensionality is higher, more creative examples are generated, and these examples less often comply with the entailment label.  

Since none of the generated datasets' accuracies is as high as the accuracy of the \textsc{OrigClass} on the original test set, we used \textsc{OrigClass} to filter the datasets subject to various prediction thresholds. 
The examples from the generated dataset were classified by \textsc{OrigClass} and if the probability of the label of the example exceeded the threshold $t \in [0.0, 0.3, 0.6, 0.9]$, then the example was retained. 

\begin{figure}	
	\begin{subfigure}{.49\textwidth}
	\centering
	\includegraphics[width=1.0\textwidth]{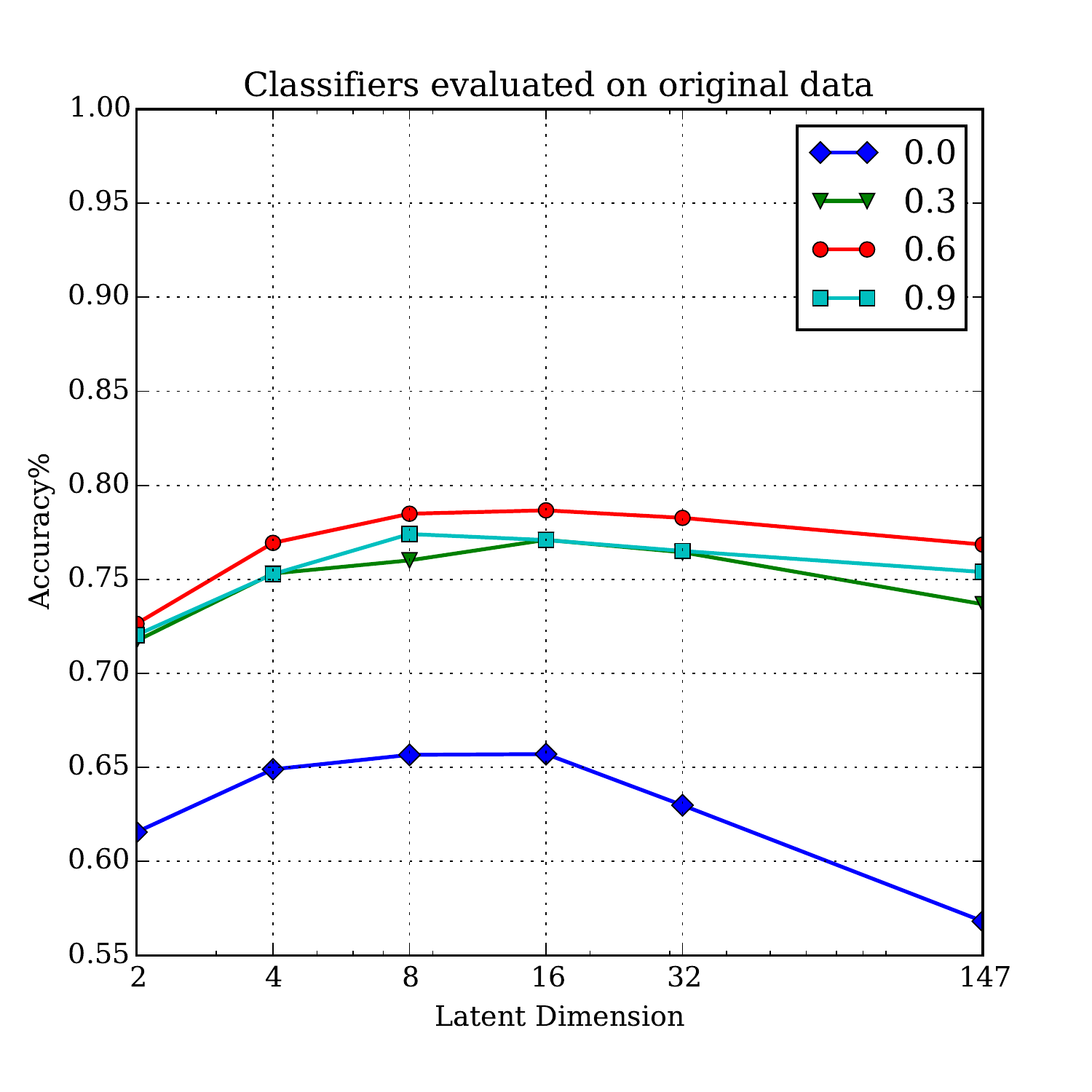}
	\caption{Accuracies of classifiers evaluated on the original test set}
	\label{fig:gen_class_acc}
	\end{subfigure}
	\begin{subfigure}{.49\textwidth}
		\centering
		\includegraphics[width=1.0\textwidth]{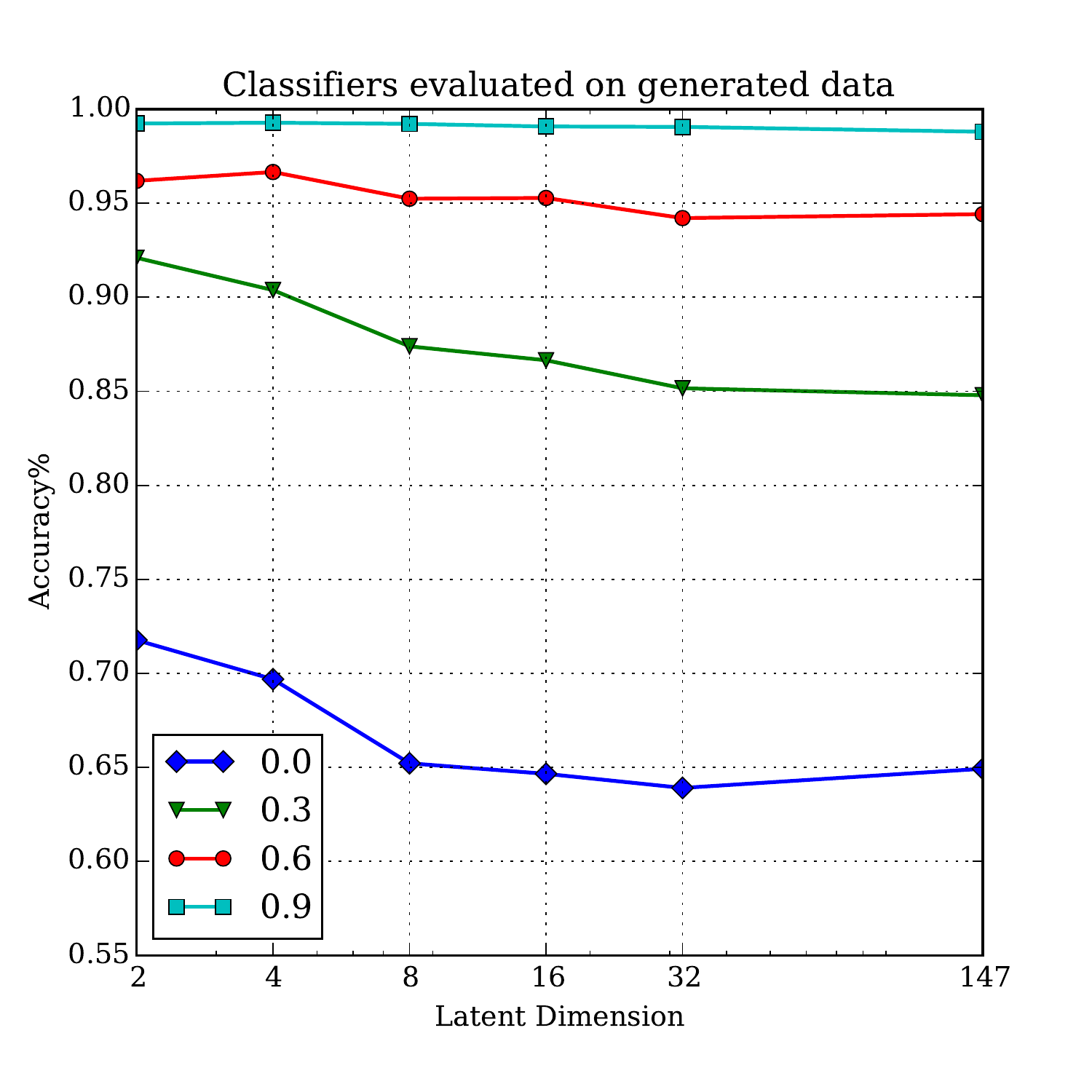}
		\caption{Accuracies of classifiers evaluated on generated development sets}
		\label{fig:gen_class_dev_acc}
	\end{subfigure}
	\caption{Accuracies of classifiers trained on the generated dataset and tested on the original test set and the generated development sets. A dataset was generated for each generative model with different latent dimension $z \in [2, 4, 8, 16, 32, 147]$. From these unfiltered datasets new datasets were created by filtering according to various prediction thresholds (0.0, 0.3, 0.6, 0.9), which also represent chart lines. A classifier was trained on each of the datasets. Each point represents the accuracy of a single classifier. The classifiers were evaluated on the original test set in Figure~\ref{fig:gen_class_acc}. Each classifier was evaluated on its corresponding generated development set in Figure~\ref{fig:gen_class_dev_acc}.}
	\label{fig:gen_class_acc_all}
	
\end{figure}

For each filtered dataset a classifier was trained. Figure~\ref{fig:gen_class_acc} shows the accuracies of these classifiers on the original test set. Filtering out the examples that have incorrect labels (according to the \textsc{OrigClass}) improves the accuracy of the classifier. However, if the threshold is set too high, the accuracy drops, since the dataset contains examples that are too trivial. Figure~\ref{fig:gen_class_dev_acc}, which represents the accuracy of classifiers on their corresponding generated development sets, further shows the trade-off between the accuracy and triviality of the examples. The classifiers trained on datasets with low latent dimension or high filtering threshold have higher accuracies. Notice that the training dataset and test dataset were generated by the same generative model.

\begin{figure}
	
	\centering
	\includegraphics[width=1.0\textwidth]{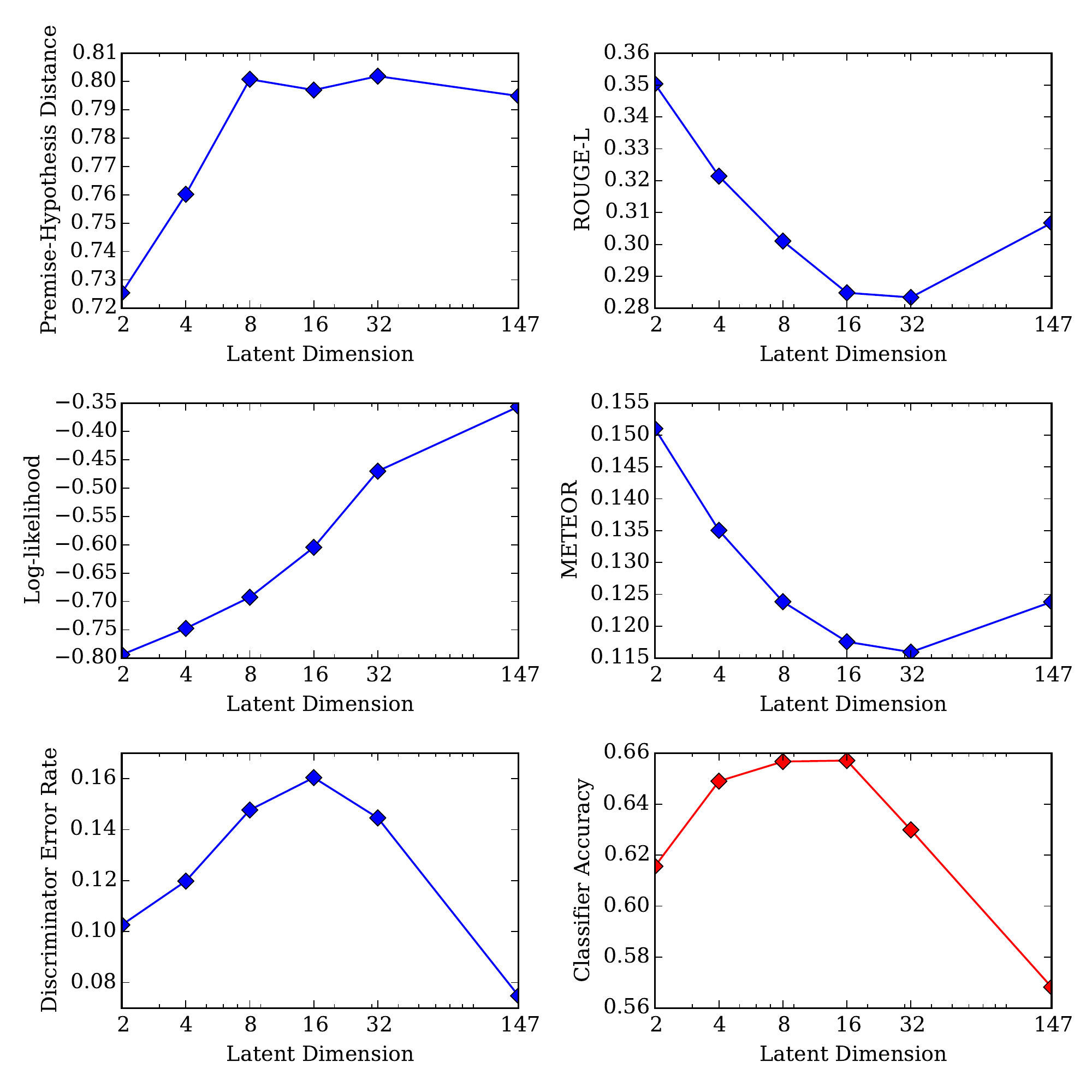}
	\caption{Comparison of unfiltered generated datasets using various metrics. Each dataset was generated by a model with a different latent dimension, then each metric was applied on each dataset. For metrics other than classifier accuracy and discriminator error rate, the metric was applied on each example and the average was calculated for each dataset.}
    
	\label{fig:other_eval}	
\end{figure}

The unfiltered datasets have been evaluated with five other metrics besides classification accuracy. The results are presented in Figure~\ref{fig:other_eval}. The whole figure shows the effect of latent dimensionality of the models on different metrics. The main purpose of the figure is not show absolute values for each of the metrics, but to compare the metrics' curves to the curve of our main metric, the accuracy of the classifier. 

The first metric -- Premise-Hypothesis Distance -- represents the average Jaccard distance between the premise and the generated hypothesis. 
Datasets generated with low latent dimensions have hypotheses more similar to premises, which indicates that the generated hypotheses are more trivial and less diverse than hypothesis generated with higher latent dimensions.

We also evaluated the models with standard language generation metrics ROUGE-L and METEOR.
The metrics are negatively correlated with the accuracy of the classifier. 
We believe this is because the two metrics reward hypotheses that are similar to their reference (original) hypothesis. However, the classifier is better if trained on more diverse hypotheses.

The next metric is the log-likelihood of hypotheses in the development set. This metric is the negative of the training loss function. The log-likelihood improves with dimensionality since it is easier to fit the hypotheses in the training step having more dimensions. Consequently, the hypothesis in the generating step are more confident -- they have lower log-likelihood.     

The last metric -- discriminative error rate -- is calculated with the discriminative model. The model is trained on the hypotheses from the unfiltered generated dataset on one side and the original hypotheses on the other side. Error rate is calculated on the (generated and original) development sets. 
Higher error rate indicates that it is more difficult for discriminative model to distinguish between the generated and the original hypotheses, which suggests that the original generating distribution and the distribution of the generative model are more similar. 
The discriminative model detects that low dimensional generative models generate more trivial examples as also indicated by the distance between premise and hypotheses. On the other hand, it also detects the hypotheses of high dimensional models, which more frequently contain grammatic or semantic errors. 

There is a positive correlation between the discriminative error rate and the accuracy of the classifier.
This observation led us to the experiment, where the generated dataset was filtered according to the prediction probability of the discriminative model. Two disjoint filtered datasets were created. One with hypotheses that had high probability that they come from the original distribution and the other one with low probability. However, the accuracies of classifiers trained on these datasets were very similar to the accuracy of the classifier on the unfiltered dataset. 
Similar test was also done with the log-likelihood metric. The examples with higher log-likelihood had similar performance than the ones with lower log-likelihood. This also lead us to set the size of the beam to 1. Also, the run time of generating hypothesis is $\mathcal{O}(b)$, where $b$ is beam size. Thus, with lower beam sizes much more hypotheses can be generated.  

To accept the hypothesis from Section~\ref{sec:nli_intro} we have shown that a quality dataset requires \emph{accurate} examples by showing that filtering the dataset with the original classifier improves the performance (Figure~\ref{fig:gen_class_acc}). Next, we have shown that \emph{non-trivial} examples are also required. If the filtering threshold is set too high, these examples are excluded, and the accuracy drops. Also, the more trivial examples are produced by low-dimensional models, which is indicated by lower premise-hypothesis distances, and lower discriminative error rate (Figure~\ref{fig:other_eval}). 
Finally, a quality dataset requires more \emph{comprehensible} examples. The high dimensional models produce less comprehensible hypotheses. They are detected by the discriminative model (see discriminator error rate in Figure~\ref{fig:other_eval}).
 
 

\subsection{Other models}

We also compared \textsc{AttEmbedDecoder} model to all other models. 
Table~\ref{tab:nli_all_models} presents the results.
For all the models the latent dimension $z$ is set to 8, as it was previously shown to be one of the best dimensions. 

For all the models the number of total parameters is relatively high, however only a portion of parameters get updated each time. The \textsc{AttEmbedDecoder} model was the best model according to our main metric -- the accuracy of the classifier trained on the generated dataset. 

The hidden dimension $d$ of the \textsc{BaseEmbedDecoder} was selected so that the model was comparable to \textsc{AttEmbedDecoder} in terms of the number of parameters $\theta_*$. 
The accuracies of classifiers generated by \textsc{BaseEmbedDecoder} are still lower than the accuracies of classifiers generated by \textsc{AttEmbedDecoder}, which shows that the attention mechanism helps the models.
\begin{table}
	\centering  	
	\begin{tabular} {lccccccccc}
	\hline
	Model & $z$ & d & $|\theta_{total}|$ &  $|\theta_*|$ & acc@0.0 & acc@0.6 & acc-data  & nll & disc-er\\
	
	\hline
	\textsc{EncoderDecoder}  & 8 & 150 & 6.4M & 1.1M &  43.4  & 72.4 & \textbf{57.5} & 1.00 & 0.01\\
	\textsc{VaeEncoderDecoder}&8 & 150 & 6.4M & 1.1M &  58.6 & 77.9 & 48.0 & 0.77 & 1.9\\
	\textsc{BaseEmbedDecoder}  & 8   & 226 & 13M  & 580K &  65.0 & 77.7 & 56.3 & 0.73 & 14.0\\
	\textsc{AttEmbedDecoder}  & 8 & 150 & 11M  & 581K &  \textbf{65.7} & \textbf{78.5} & 56.8 & \textbf{0.69} & \textbf{14.8}\\
	\hline
	\end{tabular}
	\caption{Comparison of generative models. Column $|\theta_{total}|$ is the total number of trainable parameters. Column $|\theta_*|$  represents the number of parameters that are updated with each training example. Thus, hierarchical softmax and latent representation parameters are excluded from this measure. Columns \emph{acc@0.0} and \emph{acc@0.6} represent the accuracy of the classifier trained on the unfiltered dataset and on the dataset filtered with threshold $0.6$, respectively. Column \emph{acc-data} presents the accuracy of the unfiltered development dataset evaluated by \textsc{OrigClass}. Column \emph{nll} presents the negative log-likelihood of the unfiltered development dataset. The error rates of the discriminative models are presented by \emph{disc-er}.  }
	
	\label{tab:nli_all_models}
\end{table}

Table~\ref{tab:nli_all_datasets} shows the performance of generated datasets compared to the original one.
The best generated dataset was generated by \textsc{AttEmbedDecoder}. The accuracy of its classifier is only 2.7 \% lower than the accuracy of classifier generated on the original human crafted dataset.
The comparison of the best generated dataset to the original dataset shows that the datasets had only $0.06 \%$ of identical examples. The average length of the hypothesis was $7.97$ and $8.19$ in the original dataset and in the generated dataset, respectively. 
In another experiment the generated dataset and the original dataset were merged to train a new classifier. Thus, the merged dataset contained twice as many examples as other datasets. 
The accuracy of this classifier was 82.0\%, which is 0.8 \% better than the classifier trained solely on the original training set.
However, the lowest average loss is achieved by the classifier trained on the original dataset.

\begin{table}
	\centering  	
	\begin{tabular} {lcc}
		\hline
		Dataset & loss & accuracy\%  \\
		\hline
		\textsc{EncoderDecoder}    & 1.563 & 72.4 \\
		\textsc{VaeEncoderDecoder} & 1.174 & 77.9 \\
		\textsc{BaseEmbedDecoder}  & 1.095 & 77.7  \\
		\textsc{AttEmbedDecoder}  & 0.970 & 78.5   \\
		Original Dataset & \textbf{0.475} & 81.2 \\
		Original Dataset + 	\textsc{AttEmbedDecoder} & 0.486 & \textbf{82.0}\\
		\hline
	\end{tabular}
	\caption{The performance of classifiers trained on the original and generated datasets. The classifiers were tested on original test set. The generated datasets were generated by the models from Table~\ref{tab:nli_all_models}. The generated datasets were filtered with threshold 0.6.  }
	
	\label{tab:nli_all_datasets}
\end{table}

\subsection{Qualitative evaluation}

We also did a qualitative evaluation of the generated hypothesis. Hypotheses are mostly grammatically sound. 
Sometimes the models incorrectly use indefinite articles, for instance ''\emph{an phone}'', or possessive pronouns ''\emph{a man uses her umbrella}''. These may be due to the fact the system must learn the right indefinite article for every word separately. 
On the other hand, the models sometimes generate hypotheses that showcase more advanced grammatical patterns. 
For instance, hypothesis ''\emph{The man and woman have a cake for their family}'' shows that the model can correctly use plural in a non-trivial setting. Generative neural networks have a tendency to repeat words, which sometimes make sentences meaningless, like ''\emph{A cup is drinking from a cup of coffee}'' or even ungrammatical, like ''\emph{Several people in a car car}''.

As shown previously the larger is the latent dimension more creative hypotheses are generated. 
However, with more creativity semantic errors emerge. Some hypotheses are correct, just unlikely to be written by a human, like ''\emph{A shirtless man is holding a guitar with a woman and a woman}''. Others present improbable events, like ''\emph{The girls were sitting in the park watching tv}'', or even impossible events, for instance ''\emph{The child is waiting for his wife}''. This type of errors arise because the models have not learned enough common sense logic. Finally, there are hypotheses, which make no sense. For instance, ''\emph{Two women with grassy beach has no tennis equipment}''. On the contrary, the models are able to generate some non-trivial hypotheses. 
From the original premise ''\emph{A band performing with a girl singing and a guy next to her singing as well while playing the guitar}'', the model has generated some hypotheses that do not contain concepts explicitly found in the premise. For instance, ''\emph{People are playing instruments}'' (entailment), ''\emph{The band was entirely silent}'' (contradiction), or ''\emph{The girl is playing at the concert}'' (neutral).

Regarding the compliance of the hypotheses with the label and premise, we observed that many generated hypotheses are not complying with the label, however they would be a very good example with a different label. For instance, the generated hypotheses represent entailment instead of contradiction. This also explains why the accuracy of the generated dataset measured by the original classifier is low in Figure~\ref{fig:data_acc}. On the other hand, the models generate examples that are more ambiguous and not as clear as those in the original dataset. These examples are harder to classify even for a human.
For instance, the relationship between premise ''\emph{A kid hitting a baseball in a baseball field}'' and hypothesis ''\emph{The baseball player is trying to get the ball}'' can be either interpreted either as an entailment if verb \emph{get} is intepreted as \emph{not to miss} or contradiction if \emph{get} is intepreted as \emph{possess}. 
For a deeper insight into generated hypothesis more examples are presented in \ref{sec:more_examples}.  

The gap between the discriminative error rates (disc-er) of \textsc{EncoderDecoder} models and \textsc{EmbedDecoder} models in Table~\ref{tab:nli_all_models} is significant. 
To further investigate, the same experiment was performed again by a human evaluator and the discriminative model. 
This time on a sample of 200 examples. To recap, both the model and human were asked to select the generated hypothesis given a random original and generated hypothesis without knowing which one is which.
 
Human evaluation confirms that \textsc{AttEmbedDecoder} hypotheses are more difficult to separate from the original one than the hypotheses of \textsc{VaeEncoderDecoder}.
Table~\ref{tab:discriminator} presents the results. 
The discriminative model discriminates better than the human evaluator. 
This may be due to the fact that the discriminative model has learned from a large training set, while the human was not shown any training examples.
Human evaluation has shown that generated hypotheses are positively recognized if they contain a grammatical or semantic error. 
But even if the generated hypothesis does not contain these errors, it sometimes reveals itself by not being as sophisticated as the original example. 
On the other hand, the discriminative model does not always recognize these discrepancies. 
It relies more on the differences in distributions learned form a big training set. 
The true number of non-distinguishable examples may be even higher than indicated by the human discriminator error rate since the human may have correctly guessed some of the examples he could not distinguish.

\begin{table}
	\centering  	
	
	\begin{tabular} {lcccc}
		\hline
	      & \multicolumn{2}{c}{Dev. set} & \multicolumn{2}{c}{Sample} \\
		\hline
		Gen. Model &   Disc. Model & Human & Disc. Model & Human \\
		\hline
			
		\textsc{AttEmbedDecoder}  & 14.0 & - & 14.0  & 22.5 \\
		\textsc{VaeEncoderDecoder} & 1.9 & - & 2.0 & 11.5 \\

		\hline
	\end{tabular}
	\caption{Discrimination error rate on the development set and a sample of 200 examples, evaluated by the discriminative model and human evaluator}
	
	\label{tab:discriminator}
\end{table}

\section{Conclusion}
\label{sec:nli_conclusion}
In this paper, we have proposed several generative neural networks for generating hypothesis using NLI dataset. To evaluate these models we propose the accuracy of classifier trained on the generated dataset as the main metric. The best model achieved $78.5 \%$ accuracy, which is only $2.7 \%$ less than the accuracy of the classifier trained on the original human written dataset, while the best dataset combined with the original dataset has achieved the highest accuracy. 
This model learns a decoder and a mapping embedding for each training example. 
It outperforms the more standard encoder-decoder networks.
Although more parameters are needed to be trained, less are updated on each batch.
We have also shown that the attention mechanism improves the model. 
The analysis has confirmed our hypothesis that a good dataset contains accurate, non-trivial and comprehensible examples. 
To further examine the quality of generated hypothesis, they were compared against the original human written hypotheses. The discriminative evaluation shows that in $22.5 \%$ of cases the human evaluator incorrectly distinguished between the original and the generated hypothesis. The discriminative model was actually better in distinguishing.
We have also compared the accuracy of classifier to other metrics. 
The standard text generation metrics ROUGE and METEOR do not indicate if a generated dataset is good for training a classifier. 

To obtain higher accuracies of the generated datasets, they need to be filtered, because the generative models produce examples, whose label is not always accurate. Thus, we propose for future work incorporating the classifier into the generative model, in a similar fashion that it was done on images by \citep{lamb2016discriminative}. This network could also include the discriminative model to generate examples from a distribution that is more similar to the original training distribution. 
Finally, constructing a dataset requires a lot of intensive manual work that mainly consists of writing text with some creativity.
To extend the original dataset human users could just validate or correct the generated examples.
On top of that we would like to develop active learning methods to identify incorrect generated examples that would most improve the dataset if corrected.    

\section*{Acknowledgements}
This work was supported by the Slovenian Research Agency and the ICT Programme of the EC under XLike (ICT-STREP-288342) and XLime (FP7-ICT-611346).

\section*{References}
\bibliography{nli_paper} 

\appendix

\section{More Examples}
\label{sec:more_examples}
In this section more generated hypotheses are presented. Each example starts with the original example data. Then, several hypotheses generated with from the original example with our best model are displayed. 
\vspace{0.5cm}

\small

\noindent\begin{tabular}{@{}ll}
	\textbf{Premise:} & a little boy in a blue shirt holding a toy. \\
	\textbf{Hypothesis:} & boy dressed in blue carrying toy to a toybox. \\
	\textbf{Label:}  & neutral \\
	\noalign{\vskip 0.5mm}  
	\multicolumn{2}{@{}l}{\textbf{Generated Hypotheses:}} \\
	& the little boy is caucasian. \\
	& a boy in blue holds a toy for the first time. \\
	& a child in a blue shirt holds a toy outside. \\
	& a little boy watches the toy to his friend.\\
	& a little boy is playing with a toy in the kitchen. \\
	& the boy likes to his toy. \\
	& the boy is in the store. \\

\end{tabular}

\vspace{5mm}

\noindent\begin{tabular}{@{}ll}
	\textbf{Premise:} & a little boy in a blue shirt holding a toy. \\
	\textbf{Hypothesis:} &boy dressed in red lighting things on fire. \\
	\textbf{Label:}  & contradiction \\
	\noalign{\vskip 0.5mm}  
	\multicolumn{2}{@{}l}{\textbf{Generated Hypotheses:}} \\
	& the little boy is dancing. \\
	& a boy is jumping into the pool. \\
	& a boy is watching television. \\
	& the boy is sleeping on the floor.\\
	& a boy is watching television indoors. \\
	& a boy is holding a kitten. \\
	& the little boy is standing in the pool. \\
\end{tabular}

\vspace{5mm}

\noindent\begin{tabular}{@{}ll}
	\textbf{Premise:} & a little boy in a blue shirt holding a toy. \\
	\textbf{Hypothesis:} &boy dressed in blue holds a toy.\\
	\textbf{Label:}  & entailment \\
	\noalign{\vskip 0.5mm}  
	\multicolumn{2}{@{}l}{\textbf{Generated Hypotheses:}} \\
	& the boy is holding a toy. \\
	& a kid is holding a toy.\\
	& a boy with a toy is riding a toy. \\
	& little little boy holding a toy. \\
	& a boy is riding on a toy.\\
	& a boy is with an object. \\
	& the little boy is wearing a shirt. \\
\end{tabular}

\vspace{5mm}

\noindent\begin{tabular}{@{}ll}
	\textbf{Premise:} & one woman with a dark coat on walking by herself in a city. \\
	\textbf{Hypothesis:} & the woman is walking on a sidewalk. \\
	\textbf{Label:}  & neutral \\
	\noalign{\vskip 0.5mm}  
	\multicolumn{2}{@{}l}{\textbf{Generated Hypotheses:}} \\
	& a woman at a city street has a sunny day. \\
	& the woman is shopping.\\
	& a woman is walking to the store. \\
	& the woman is asian. \\
	& the lady with the coat walking by her way to a store in the city.\\
	& the woman is afraid of the woman. \\
	& a woman was watching to go to a party \\
\end{tabular}

\vspace{5mm}

\noindent\begin{tabular}{@{}ll}
	\textbf{Premise:} & one woman with a dark coat on walking by herself in a city. \\
	\textbf{Hypothesis:} & the woman is standing in a building. \\
	\textbf{Label:}  & contradiction \\
	\noalign{\vskip 0.5mm}  
	\multicolumn{2}{@{}l}{\textbf{Generated Hypotheses:}} \\
	& the woman is sitting on a couch. \\
	& a woman is sitting on the ground.\\
	& a woman in a coat and a coat is sitting next to her way \\
	& a woman is watching the movie at the theater. \\
	& a man is walking on the street\\
	& the men are asleep \\
	& the dog is wearing a pink coat. \\
\end{tabular}

\vspace{5mm}

\noindent\begin{tabular}{@{}ll}
	\textbf{Premise:} & one woman with a dark coat on walking by herself in a city. \\
	\textbf{Hypothesis:} & the woman is outdoors. \\
	\textbf{Label:}  & entailment \\
	\noalign{\vskip 0.5mm}  
	\multicolumn{2}{@{}l}{\textbf{Generated Hypotheses:}} \\
	& a female is walking \\
	& there are people present.\\
	& woman walk. \\
	& a woman in a coat walking through a city.\\
	& there is a woman and a man in a crosswalk.\\
	& woman are walking by a city. \\
	& the woman is talking to each other. \\
\end{tabular}

\vspace{5mm}
\newdimen\origiwspc
\origiwspc=\fontdimen2\font
\noindent\begin{tabular}{@{}ll}
	\textbf{Premise:} & \fontdimen2\font=2pt 
	   two young boys of opposing teams play football, while wearing full protection uniforms and helmets.\fontdimen2\font=\origiwspc \\
	\textbf{Hypothesis:} & boys scoring a touchdown \\
	\textbf{Label:}  & neutral \\
	\noalign{\vskip 0.5mm}  
	\multicolumn{2}{@{}l}{\textbf{Generated Hypotheses:}} \\
	& the boys in uniform are trying to win the game. \\
	& the brothers are wearing a uniforms for the football game. \\
	& the boys were waiting to play the game. \\
	& two boys are fighting with their team. \\
	& two teams wearing uniforms being finished for the game \\
	& two young boys are having fun. \\
	& the boys are wearing helmets and are standing nearby.  \\
\end{tabular}

\vspace{5mm}

\noindent\begin{tabular}{@{}ll}
	\textbf{Premise:} & \fontdimen2\font=2pt 
	two young boys of opposing teams play football, while wearing full protection uniforms and helmets. \fontdimen2\font=\origiwspc\\
	\textbf{Hypothesis:} & dog eats out of bowl \\
	\textbf{Label:}  & contradiction \\
	\noalign{\vskip 0.5mm}  
	\multicolumn{2}{@{}l}{\textbf{Generated Hypotheses:}} \\
	& the two boys are watching tv. \\
	& the boys are playing baseball. \\
	& the man is on a trampoline. \\
	& two young girls are dressed in tuxedos. \\
	& two kids playing with the ball at the beach. \\
	& two kids are in the ocean. \\
	& the boys are chasing an elephant  \\
\end{tabular}
	
\vspace{5mm}

\noindent\begin{tabular}{@{}ll}
	\textbf{Premise:} & \fontdimen2\font=2pt 
	two young boys of opposing teams play football, while wearing full protection uniforms and helmets. \fontdimen2\font=\origiwspc\\
	\textbf{Hypothesis:} & boys play football \\
	\textbf{Label:}  & entailment \\
	\noalign{\vskip 0.5mm}  
	\multicolumn{2}{@{}l}{\textbf{Generated Hypotheses:}} \\
	& a trio of young boys wearing uniforms. \\
	& two kids are being photographed for a football player. \\
	& the boys have uniforms. \\
	& two boys are wearing uniforms. \\
	& the boys are dressed \\
	& two boys are performing. \\
	& the boys all in uniforms for their team.\\
	
\end{tabular}

\end{document}